\documentclass[11pt]{article}

\usepackage[final]{acl}

\makeatletter
\renewcommand{\outauthor}{
    \begin{tabular}[t]{c}
        \bf\@author
    \end{tabular}
}
\makeatother
\usepackage{amsmath}
\usepackage{amsfonts}
\usepackage{amsmath}
\usepackage{times}
\usepackage{latexsym}

\usepackage[T1]{fontenc}

\usepackage[utf8]{inputenc}

\usepackage{microtype}

\usepackage{inconsolata}

\usepackage{graphicx}

\usepackage{amsmath}
\usepackage{longtable}  
\usepackage{booktabs}   

\title{DEO: Training-Free Direct Embedding Optimization for\\ Negation-Aware Retrieval}

\author{
  \textbf{Taegyeong Lee}$^{1}$\thanks{Equal contribution} \quad 
  \textbf{Jiwon Park}$^{2}$\footnotemark[1] \quad 
  \textbf{Seunghyun Hwang}$^{3}$\footnotemark[1] \quad 
  \textbf{JooYoung Jang}$^{1}$\thanks{Corresponding author} \\
  \addlinespace[0.2cm]
  $^{1}$Miri.DIH \quad $^{2}$Department of Industrial Engineering, Hanyang University \\
  $^{3}$Department of Applied Data Science, Sungkyunkwan University \\
  \addlinespace[0.2cm]
  \texttt{\small tglee@miridih.com, jiwonpark@hanyang.ac.kr,} \\
  \texttt{\small hsh1030@g.skku.edu, jyjang@miridih.com}
}

\begin{document}
\maketitle

\begin{abstract}
Recent advances in Large Language Models (LLMs) and Retrieval-Augmented Generation (RAG) have enabled diverse retrieval methods. However, existing retrieval methods often fail to accurately retrieve results for negation and exclusion queries. To address this limitation, prior approaches rely on embedding adaptation or fine-tuning, which introduce additional computational cost and deployment complexity. We propose \emph{\textbf{Direct Embedding Optimization (DEO)}}, a training-free method for negation-aware text and multimodal retrieval. DEO decomposes queries into positive and negative components and optimizes the query embedding with a contrastive objective. Without additional training data or model updates, DEO outperforms baselines on NegConstraint, with gains of +0.0738 nDCG@10 and +0.1028 MAP@100, while improving Recall@5 by +6\% over OpenAI CLIP in multimodal retrieval. These results demonstrate the practicality of DEO for negation- and exclusion-aware retrieval in real-world settings. The code is publicly available at 
\href{https://github.com/taegyeong-lee/DEO-negation-aware-retrieval}{GitHub}.
\end{abstract}
    
\section{Introduction}
Recent advances in Large Language Models (LLMs)~\cite{grattafiori2024llama,yang2025qwen3,team2024qwen2} and Retrieval-Augmented Generation (RAG)~\cite{lewis2020retrieval,xu2025logical,zeighami2024nudge} have enabled systems to generate responses conditioned on user inputs and aligned with user intent. However, real-world queries frequently include negation and exclusion, posing challenges for consistently retrieving content that accurately reflects user intent~\cite{cai2025tng,singh2023nlms,dong2023towards}.

Existing approaches~\cite{zeighami2024nudge, shevkunovrelevance,wang2024improving} have attempted to improve retrieval performance through fine-tuning embedding models. While effective, these approaches require substantial GPU resources, large-scale fine-tuning datasets, and extensive training, which restrict their applicability in resource-constrained environments. In addition, such methods can potentially degrade retrieval performance and lack clear controllability, particularly in handling queries involving negation and exclusion. Consequently, effectively capturing user intent in retrieval under queries with negation and exclusion remains both challenging and important~\cite{singh2023nlms, alhamoud2025vision}.

Recently, a method has leveraged sparse autoencoders (SAE)~\cite{kang2025interpret} to interpret and control dense embeddings through latent sparse features, while maintaining comparable retrieval accuracy. NUDGE~\cite{zeighami2024nudge}, on the other hand, proposes a non-parametric embedding fine-tuning approach that directly modifies embeddings of data records to maximize k-NN retrieval accuracy, achieving significant improvements in both performance and efficiency over full model fine-tuning and adaptor-based methods. However, these methods still rely on fine-tuning and thus require large datasets and substantial GPU resources, limiting their practicality.

Therefore, we propose \emph{\textbf{Direct Embedding Optimization (DEO)}}, a simple yet effective approach for multimodal retrieval that does not require fine-tuning. First, DEO decomposes user queries into positive and negative sub-queries using LLMs. For example, given a query such as ``show me the latest earnings forecast, but exclude 2024 results,'' the LLM generates positive queries (e.g., ``earnings forecast for 2025,'' ``financial statements'') and negative queries (e.g., ``2024 earnings,'' ``2024 financial report''), thereby explicitly separating user intent into inclusion and exclusion components.

Second, both the original user query and the decomposed positive and negative sub-queries are embedded using a pre-trained embedding model. We directly optimize the user query embedding space via a contrastive loss, pulling it closer to the positive query embeddings and pushing it away from the negative query embeddings. This optimization aligns the original query embedding with more specific and semantically rich positive queries, while simultaneously distancing it from negative queries, thereby producing embeddings that better represent user intent involving negation and exclusion.

Finally, retrieval is performed using the optimized embeddings, enabling improved performance without additional fine-tuning.

Our experiments show that DEO improves MAP on the NegConstraint benchmark from 0.6299 to 0.7327 and nDCG@10 from 0.7139 to 0.7877 in the best-performing configuration, using the BGE-large-en-v1.5 embedding model. On multimodal retrieval (COCO-Neg)~\cite{alhamoud2025vision}, DEO notably increases Recall@5 with OpenAI CLIP~\cite{radford2021learning} from 0.4792 to 0.5392. Moreover, DEO consistently improves performance across all evaluated text retrieval benchmarks. These results demonstrate that our method provides stable absolute gains, making it robust for real-world retrieval settings and effective in handling negation and exclusion queries.

Our main contributions are as follows:

\begin{itemize}
 \item We propose Direct Embedding Optimization (DEO), an effective retrieval method without fine-tuning or additional datasets.
\item By directly optimizing the embedding space via contrastive loss over positive and negative sub-queries, DEO enables negation- and exclusion-aware retrieval that more precisely captures user intent.
 \item DEO is model- and modality-agnostic, generalizing across diverse embedding models and retrieval settings, and experiments demonstrate consistent improvements over baselines on both text and multimodal benchmarks. 
\end{itemize}

\section{Related Work}

\subsection{Dense Retrieval}
Dense retrieval encodes queries and documents into continuous, low-dimensional embeddings that capture semantic similarity, and typically outperforms sparse term-matching methods based on high-dimensional vectors \citep{kang2025interpret,karpukhin2020dense}. Prior research has explored improvements through training strategies, distillation, and pre-training. Transfer learning on large-scale datasets such as MS MARCO~\cite{bajaj2016ms} has also been widely adopted, though it is resource-intensive to construct \citep{singh2023nlms}. More recently, zero-shot dense retrieval has reduced the reliance on explicit relevance labels. Despite these advances, the problem of effectively handling queries with negation and exclusion remains relatively underexplored.

\subsection{Embedding Control and Fine-tuning Alternatives}
Beyond full model fine-tuning, recent work has explored methods for directly controlling or refining embedding space to improve retrieval. Representative approaches include projecting dense embeddings into interpretable sparse latent features and applying non-parametric optimization to directly adjust record embeddings for improved k-NN accuracy \citep{zeighami2024nudge,shevkunovrelevance,wang2024improving}. While these methods can be effective, they often require sizable datasets and substantial GPU resources, which limits their applicability in resource-constrained settings. This motivates the development of lightweight approaches that better capture nuanced aspects of user intent—particularly negation and exclusion—without additional fine-tuning.

\begin{figure*}[t!] 
\includegraphics[width=\textwidth]{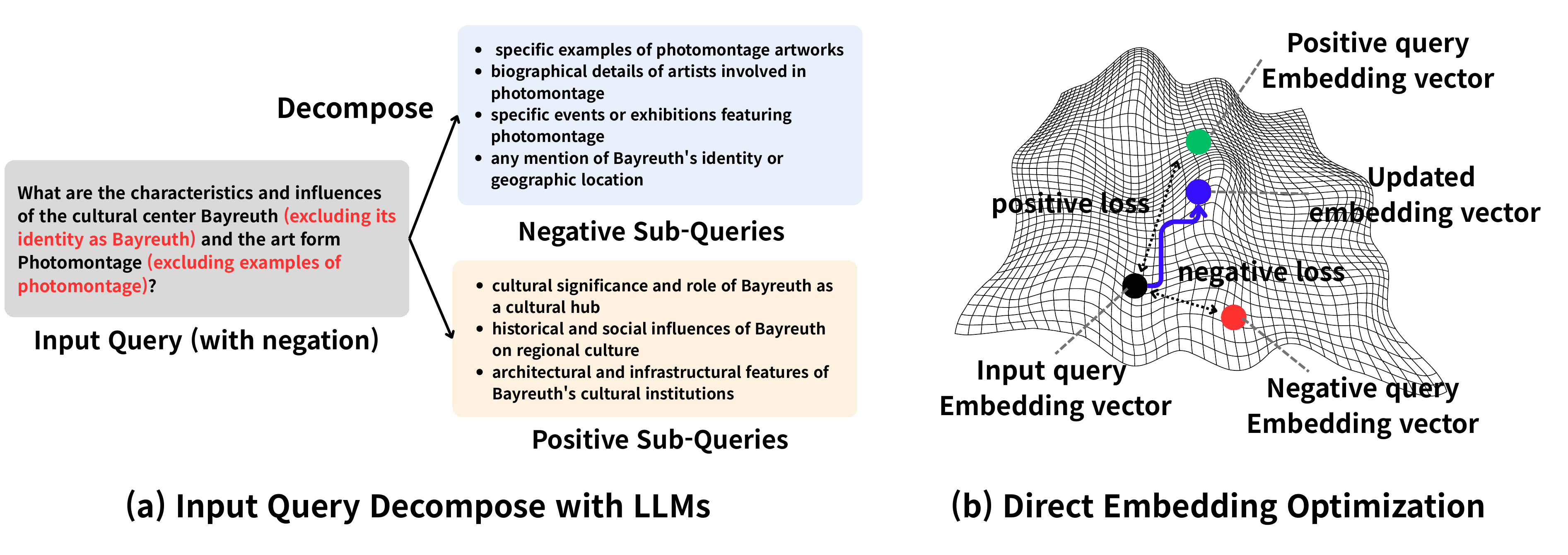}
\caption{\textbf{Overview of the proposed Direct Embedding Optimization (DEO).} 
(a) Given an input query containing negation, we use an LLM to decompose it into positive and negative sub-queries. 
(b) The input query embedding is then optimized with a contrastive loss by pulling it closer to positive query embeddings and pushing it farther from negative query embeddings, enabling negation- and exclusion-aware retrieval.}
\label{figure_overview}
\end{figure*}

\subsection{Negation- and Exclusion-aware Retrieval}
Queries containing negation or exclusion require an explicit distinction between inclusion and exclusion semantics, which standard retrievers often fail to capture reliably \citep{singh2023nlms,alhamoud2025vision}. Prior work has attempted to address this issue through fine-tuning or task-specific regularization to improve sensitivity to negation \citep{wang2024improving,zeighami2024nudge}, but such approaches typically incur substantial computational cost and offer limited controllability. An alternative direction is to leverage large language models to decompose queries into semantically coherent sub-queries, thereby providing explicit representations of user intent. Building on this idea, our method directly optimizes query embeddings with respect to positive and negative sub-queries using a contrastive objective, which improves alignment with user intent while avoiding model fine-tuning.

\subsection{Negation Robustness in Multimodal Retrieval}
\label{subsec:mm-neg}
Vision–language retrievers trained with large-scale contrastive learning, such as CLIP~\cite{radford2021learning}, learn a shared image–text embedding space and achieve strong zero-shot retrieval performance. BLIP~\cite{li2022blip} and BLIP-2~\cite{li2023blip} extend these capabilities to captioning and VQA (Visual Question Answering)~\cite{antol2015vqa, kim2021visual} while maintaining competitive retrieval accuracy. However, despite strong average performance, these models remain brittle to negation phenomena, including attribute negation (e.g., “not red”), absence (e.g., “no person”), and relational negation (e.g., “A is not left of B”). NegBench~\citep{alhamoud2025vision} evaluates this limitation through retrieval and multiple-choice tasks with negated captions across image and video domains, confirming that existing models struggle to distinguish affirmative from negated statements.

In contrast, our approach enables negation- and exclusion-aware retrieval across both text and multimodal settings by providing explicit control over semantic components relevant to truth conditions, and demonstrates robustness under established evaluation benchmarks.
\vspace{-0.3cm}
\section{Method}
\label{sec:method}
We propose a simple yet effective retrieval method. As illustrated in Figure~\ref{figure_overview}, our model consists of two stages: (a) Decomposing the user query into positive and negative sub-queries. (b) Directly optimizing the embedding space of input query as a parameter by using contrastive loss. Through this approach, we enable negation- and exclusion-aware retrieval in the embedding space without fine-tuning embedding models.

\subsection{Query Decomposition}
In real-world scenarios, user queries frequently contain negation or exclusion, expressed using phrases such as ``exclude'' or ``do not mention.'' To address this challenge, as shown in Figure~\ref{figure_overview} (a), we employ a large language model (LLM) in a prompt-based setting to semantically analyze the input query and explicitly capture its negation or exclusion intent. The LLM then decomposes the original query into structured positive and negative sub-queries.

For example, given the query ``What are the characteristics and influences of the cultural center Bayreuth (excluding its identity as Bayreuth) and the art form Photomontage (excluding examples of photomontage)?'', our method separates the retrieval-relevant components from the exclusion constraints and generates corresponding positive and negative sub-queries. These sub-queries are subsequently embedded and optimized independently, yielding a structured representation that more accurately reflects the user’s intent under negation and exclusion.

The positive sub-queries are an enriched version of the user’s request: 
[``cultural significance and role of Bayreuth as a cultural hub'', ``historical and social influences of Bayreuth on regional culture'', ``architectural and infrastructural features of Bayreuth's cultural institutions''].
This expansion captures the essence of the user’s intent while expressing it in a more elaborate and comprehensive form. 

The negative sub-queries explicitly encode the exclusionary intent: [``specific examples of photomontage artworks'', ``biographical details of artists involved in photomontage'', ``specific events or exhibitions featuring photomontage'', ``any mention of Bayreuth's identity or geographic location'']. The negation is semantically expanded so that the system can clearly identify and filter out the unwanted element.

Through this structured decomposition, the original query becomes both clarified and operationalized: the positive sub-queries guide semantic relevance expansion, whereas the negative sub-queries impose explicit exclusion constraints during retrieval.

\subsection{Direct Embedding Optimization}
Unlike previous methods~\cite{zeighami2024nudge,patel2024tripletclip,kang2025interpret} that require fine-tuning or additional supervised datasets, we directly optimize the embedding of the input query at inference time while keeping the encoder frozen. Let $E
(\cdot)$ denote the encoder. Given an input query $q$, we obtain its original embedding:  
  
\begin{equation}
\small
\mathbf{e}_o = E(q) \in \mathbb{R}^d
\end{equation}

From the LLM-based query decomposition, we obtain a set of positive sub-queries $P=\{p_i\}_{i=1}^{K}$ and a set of negative sub-queries  $N=\{n_j\}_{j=1}^{M}$. Their embeddings are computed as: 

\begin{equation}
\small
\begin{aligned}
\mathbf{e}_{p_i} = E(p_i), \quad
\mathbf{e}_{n_j} = E(n_j).  
\end{aligned}
\end{equation}

We initialize a learnable query embedding $\mathbf{e}_u$ with the original embedding.

\begin{equation}
\small
\begin{aligned}
\mathbf{e}_u \leftarrow \mathbf{e}_o.
\end{aligned}
\end{equation}

The objective consists of three components:
(i) an attraction term that pulls the optimized embedding toward the positive embeddings,
(ii) a repulsion term that pushes it away from the negative embeddings, and
(iii) a consistency term that preserves the semantics of the original query.  The loss function is defined as:
\begin{equation}
\small
\begin{aligned}
\mathcal{L}(\mathbf{e}_u) =
&\lambda_p \cdot \frac{1}{K} \sum_{i=1}^{K}
\| \mathbf{e}_u - \mathbf{e}_{p_i} \|^2 \\
&- \lambda_n \cdot \frac{1}{M} \sum_{j=1}^{M}
\| \mathbf{e}_u - \mathbf{e}_{n_j} \|^2 \\
&+ \lambda_o \cdot \| \mathbf{e}_u - \mathbf{e}_o \|^2.
\end{aligned}
\end{equation}

Here, $\lambda_p$, $\lambda_n$, and $\lambda_o$ are hyperparameters controlling the strength of positive attraction, negative repulsion, and consistency regularization, respectively.

We minimize $\mathcal{L}$ using a gradient-based optimizer (Adam) for a fixed number of steps while keeping the encoder parameters unchanged. The resulting optimized embedding $\mathbf{e}_u$ is then used as the final query representation for retrieval.
 
\subsection{Retrieval with Optimized Embeddings}
We derive an optimized embedding space that both enriches the user query and incorporates its negation or exclusion intent. Retrieval is then performed directly with this embedding. To further emphasize exclusion or restrict specific documents, the retrieval behavior can be adjusted by tuning the optimization hyperparameters $\lambda_p$ and $\lambda_n$, which weight the contributions of positive and negative sub-queries. Importantly, our approach is model-agnostic: it can be applied to various embedding models, including multimodal encoders such as CLIP~\cite{radford2021learning}, and enables effective multimodal retrieval without additional fine-tuning or task-specific datasets.
\section{Experiments}
\subsection{Experimental Setup}
\label{sec:exp_setup}
\textbf{Datasets and Metrics.} 
We primarily focus on the task of negation-aware retrieval. For this, we employ the NegConstraint~\cite{xu2025logical} and NevIR~\cite{weller2024nevir} datasets. Following prior work, we evaluate performance on NegConstraint using nDCG@10 and MAP@100, while NevIR is evaluated using the Pairwise metric.
In addition, for negation-aware text-to-image retrieval, we evaluate on a text-to-image retrieval dataset that contains negation, namely the COCO-Neg from NegBench~\cite{alhamoud2025vision}, using Recall@5 as the evaluation metric.

\noindent\textbf{Baselines.} 
Because our approach does not require fine-tuning or additional training data, it can be directly applied to any embedding model.
For text retrieval, we compare against BGE-M3, BGE-large-en-v1.5, and BGE-small-en-v1.5.
For text-to-image retrieval, we consider OpenAI CLIP~\cite{radford2021learning}, CLIP-laion400m, CLIP-datacomp, and NegCLIP~\cite{yuksekgonul2022and} as baselines.

\noindent\textbf{Implementation Details.} 
We used the [CLS] token representation for all embedding models, employed the FAISS library, and computed cosine similarity for retrieval. 
For query decomposition in our method, we used OpenAI’s GPT-4.1-nano with a temperature of 0.1. Across all text-only datasets, we set the hyperparameters $\lambda_p = 1$, $\lambda_n = 1$, and $\lambda_o = 0.2$, and use 20 optimization steps. For multimodal retrieval datasets, we instead set $\lambda_p = 1.0$, $\lambda_n = 1.0$, and $\lambda_o = 1.0$. 

\begin{table}[]
\small
\centering
\begin{tabular}{lccc}
\hline
                  & \multicolumn{2}{c}{NegConstraint} & \multicolumn{1}{c}{NevIR}    \\
                  & MAP             & nDCG& \multicolumn{1}{l}{Pairwise} \\ \hline
BGE-small-en-v1.5 & 0.6702         & 0.7372          & 0.1569                       \\
w/ DEO            & \textbf{0.7302} & \textbf{0.7795} & \textbf{0.1894}                       \\
BGE-large-en-v1.5 & 0.6299          & 0.7139          & 0.2552                       \\
w/ DEO            & \textbf{0.7327} & \textbf{0.7877} & \textbf{0.2776 }                      \\
BGE-M3            & 0.6374         & 0.7250          & 0.2668                       \\
w/ DEO            & \textbf{0.7379} & \textbf{0.7946} &\textbf{0.2928}                      \\ \hline
\end{tabular}
\caption{\textbf{Performance on the NegConstraint~\cite{xu2025logical} benchmark for negation retrieval.} ``w/ DEO'' represents the performance after applying our proposed method. All nDCG scores correspond to nDCG@10.}
\label{table:neg_text}
\end{table}

\begin{table}[]
\small
\centering
\begin{tabular}{lc}
\hline
               & Recall@5        \\ \hline
OpenAI CLIP    & 0.4792         \\
w/ DEO         & \textbf{0.5392} \\
CLIP-laion400m & 0.5248          \\
w/ DEO         & \textbf{0.5737} \\
CLIP-datacomp  & 0.4984         \\
w/ DEO         & \textbf{0.5513} \\
NegCLIP        & 0.6715         \\
w/ DEO         & \textbf{0.6980} \\ \hline
\end{tabular}
\caption{\textbf{Performance on the COCO-Neg benchmark for text-to-image negation retrieval~\cite{alhamoud2025vision}}.  ``w/ DEO'' represents the performance after applying our proposed method.}
\label{table:neg_image}
\vspace{-0.4cm}
\end{table}

\subsection{Main Results}
\subsubsection{Negation Benchmark}

We evaluate retrieval performance on negation- and exclusion-intensive tasks using the NegConstraint and NevIR benchmarks. As shown in Table~\ref{table:neg_text}, applying DEO consistently improves performance across all BGE variants, yielding gains in MAP, nDCG@10, and pairwise accuracy.

Across all models, DEO leads to clear improvements on NegConstraint, indicating better retrieval under negation constraints. For example, on BGE-large-en-v1.5, applying DEO increases MAP by +0.1028 (+16.32\%) and nDCG@10 by +0.0738 (+10.34\%). These gains are further supported by results on NevIR, which directly evaluates negation-aware pairwise discrimination. In particular, DEO improves the NevIR pairwise score from 0.1569 to 0.1894 for BGE-small-en-v1.5, from 0.2552 to 0.2776 for BGE-large-en-v1.5, and from 0.2668 to 0.2928 for BGE-M3.

The consistent improvements across both ranking-based metrics and pairwise negation evaluation suggest that DEO enhances the model’s ability to correctly interpret and retrieve under negated query conditions. Overall, these results demonstrate that DEO improves negation-sensitive retrieval performance across diverse embedding models.

\subsubsection{Negation Benchmark on Text-to-Image Retrieval}
Our method is applicable regardless of the embedding model or modality. To verify this, we evaluate on COCO-Neg~\cite{alhamoud2025vision}. 

As shown in Table~\ref{table:neg_image}, DEO consistently improves Recall@5 across all four CLIP-based models. OpenAI CLIP shows the largest gain, with Recall@5 increasing by 6.00\% over the baseline, and CLIP-datacomp and CLIP-laion400m also achieve improvements of 5.29\% and 4.89\%,
respectively. Notably, even NegCLIP, which is explicitly fine-tuned for compositional understanding, improves from 0.6715 to 0.6980, confirming that DEO yields additional gains even on top of negation-aware models. These results confirm that our method generalizes effectively to multimodal retrieval settings.

\subsection{Ablation Study}
\label{sec:ablation}
To analyze the contribution of each component, we conduct ablation studies from 4 perspectives: the choice of LLM, balancing parameters ($\lambda_o$, $\lambda_n$, $\lambda_p$), the effect of query decomposition, and the optimization procedure.

\begin{table}[]
\small
\centering
\begin{tabular}{lcc}
\hline
                           & \multicolumn{2}{c}{NegConstraint} \\
\multicolumn{1}{c}{Method} & MAP             & nDCG@10            \\ \hline
BGE-S w/ Qwen              & 0.7144          & 0.7705          \\
BGE-S w/ GPT               & \textbf{0.7302} & \textbf{0.7795} \\
BGE-L w/ Qwen              & 0.6859          & 0.7553          \\
BGE-L w/ GPT               & \textbf{0.7327} & \textbf{0.7877} \\
BGE-M3 w/ Qwen             & 0.7280          & 0.7871          \\
BGE-M3 w/ GPT              & \textbf{0.7379} & \textbf{0.7946} \\ \hline
\end{tabular}
\caption{\textbf{Performance comparison on the NegConstraint dataset using different LLM backbones for query decomposition.} 
Qwen refers to Qwen2.5-1.5B-Instruct, and GPT refers to GPT-4.1-nano. 
BGE-S, BGE-L denote BGE-small-en-v1.5, BGE-large-en-v1.5, respectively.}
\vspace{-0.3cm}
\label{table:qwen_text}
\end{table}

\begin{table}[]
\small
\centering
\begin{tabular}{lc}
\hline
                           & COCO-Neg        \\
\multicolumn{1}{c}{Method} & Recall@5        \\ \hline
CLIP-laion400m w/ Qwen               & 0.5656          \\
CLIP-laion400m w/ GPT                & \textbf{0.5737} \\
NegCLIP w/ Qwen            & 0.6872          \\
NegCLIP w/ GPT             & \textbf{0.6980} \\ \hline
\end{tabular}
\caption{\textbf{Performance comparison on COCO-Neg using different LLM backbones for query decomposition.}}
\label{table:qwen_image}
\vspace{-0.4cm}
\end{table}

\begin{table}[t]
\small
\centering
\begin{tabular}{lccc}
\hline
                               & \multicolumn{2}{c}{NegConstraint} & \multicolumn{1}{l}{COCO-Neg} \\
\multicolumn{1}{c}{Parameters} & MAP             & nDCG@10         & \multicolumn{1}{c}{Recall@5}  \\ \hline
Baseline             & 0.6374      & 0.7250        & 0.4792                  \\ \hline
0.2, 1.0, 1.0                  & \textbf{0.7379}        & \textbf{0.7946}         & 0.5349                      \\
0.2, 1.0, 2.0                  & 0.7366          & 0.7942        & 0.5275                      \\
0.2, 2.0, 1.0                  & 0.7160 & 0.7769 & 0.5237             \\
1.0, 1.0, 1.0                  & 0.6713          & 0.7510         & \textbf{0.5392}                     \\
1.0, 1.0, 2.0                  & 0.7034         & 0.7740          & 0.5303                      \\
1.0, 2.0, 1.0                  & 0.7013        & 0.7712        & 0.5246                       \\ \hline
\end{tabular}
\caption{\textbf{Effect of varying the weight balancing parameters $\lambda_o$ , $\lambda_p$, and $\lambda_n$ on the NegConstraint and COCO-Neg.}
NegConstraint results use BGE-M3, and COCO-Neg results use OpenAI CLIP. Optimization step is 20.}
\label{table:weight_balance}
\end{table}

\begin{table}[t!]
\small
\centering
\begin{tabular}{lcc}
\hline
                           & \multicolumn{2}{c}{NegConstraint} \\
\multicolumn{1}{c}{Method} & MAP             & nDCG@10         \\ \hline
BGE-M3                     & 0.6374         & 0.7250        \\
Ours (only decompose, AVG)      & 0.6451 & 0.7312 \\
Ours (only decompose, RRF)  & 0.6641          & 0.7417          \\
Ours (full)                & \textbf{0.7379} & \textbf{0.7946} \\ \hline
\end{tabular}
\caption{\textbf{Ablation results on NegConstraint with BGE-M3.} Query decomposition alone yields limited gains, while most performance improvements arise from embedding optimization rather than decomposition itself.}
\label{ablation_decompose}
\vspace{-0.3cm}
\end{table}

\subsubsection{Effect of Different LLMs}
To evaluate how an LLM's decomposition ability impacts our method, we conduct experiments with different backbone models. Specifically, we compare GPT-4.1-nano against Qwen2.5-1.5B-Instruct~\cite{team2024qwen2}, which has far fewer parameters. As shown in Tables~\ref{table:qwen_text} and~\ref{table:qwen_image}, GPT-4.1-nano consistently outperforms Qwen2.5-1.5B-Instruct across all embedding models on both NegConstraint and COCO-Neg, which we attribute to more precise query decompositions from the larger model. Nevertheless, even with Qwen2.5-1.5B-Instruct, our method achieves notable improvements over the baselines, indicating that DEO delivers consistent gains regardless of LLM scale.

\subsubsection{Effect of Weight Balancing}
We analyze the impact of the weight balancing parameters $\lambda_{o}$ (original consistency), $\lambda_{p}$ (positive), and $\lambda_{n}$ (negative) across both the NegConstraint and COCO-Neg benchmarks. As shown in Table \ref{table:weight_balance}, DEO consistently outperforms the baseline across all tested configurations, demonstrating its inherent robustness to hyperparameter variations. On the text-only NegConstraint dataset, the best configuration reached a MAP of 0.7379 and an nDCG@10 of 0.7946 by setting $\lambda_{o} = 0.2$. 

In contrast, COCO-Neg peaked at 0.5392 Recall@5 with $\lambda_{o} = 1.0$, as preserving the original semantic context is essential to maintain alignment in shared vision-language spaces like CLIP. These findings indicate that DEO generalizes effectively across diverse modalities through its adaptive optimization objective.

\begin{figure}[t!] 
\includegraphics[width=\linewidth]{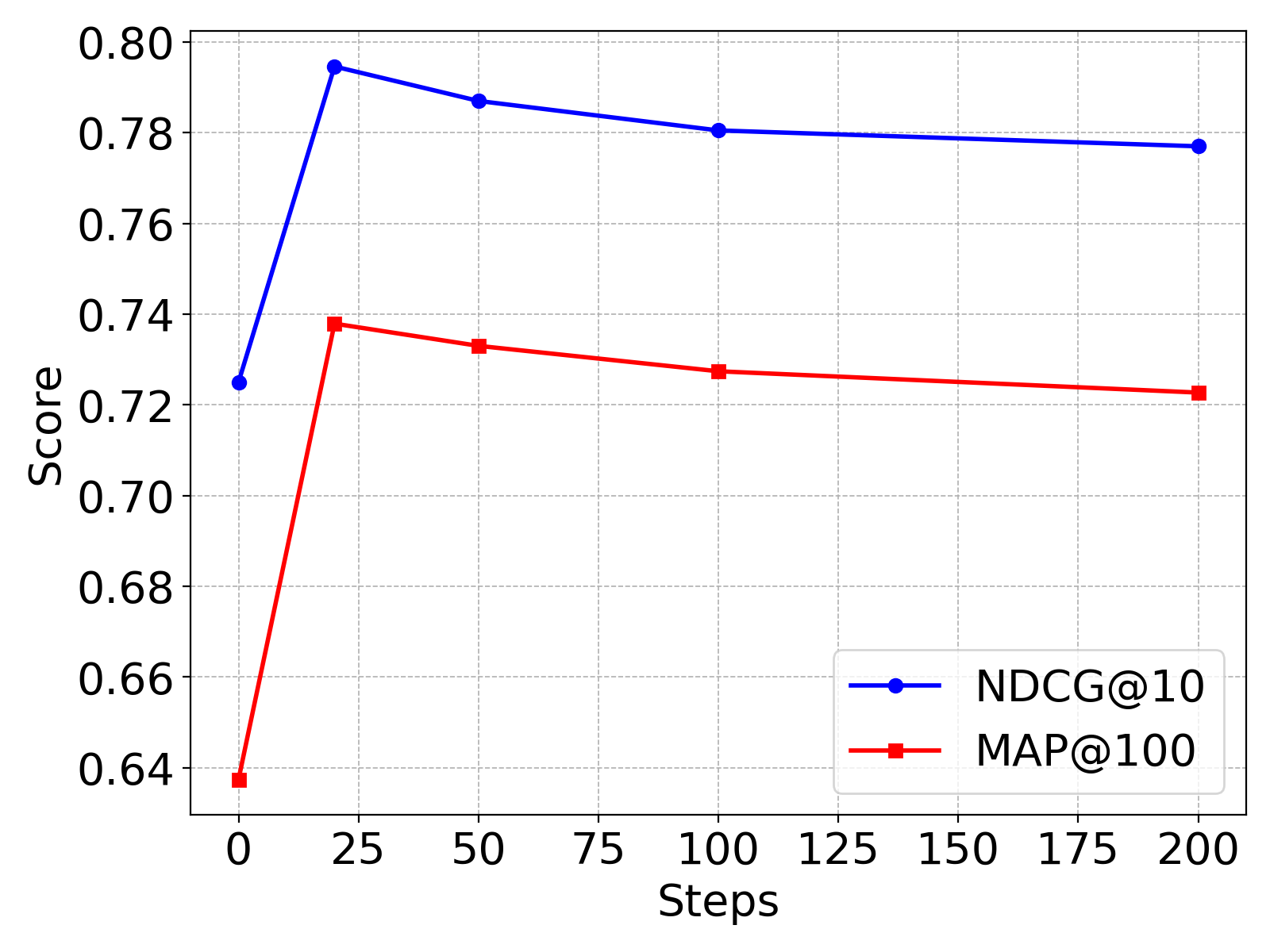}
    \caption{\textbf{Retrieval performance with respect to the number of optimization steps on NegConstraint}.}
\label{figure:opt_text}
\end{figure}

\begin{figure}[t!] 
\includegraphics[width=\linewidth]{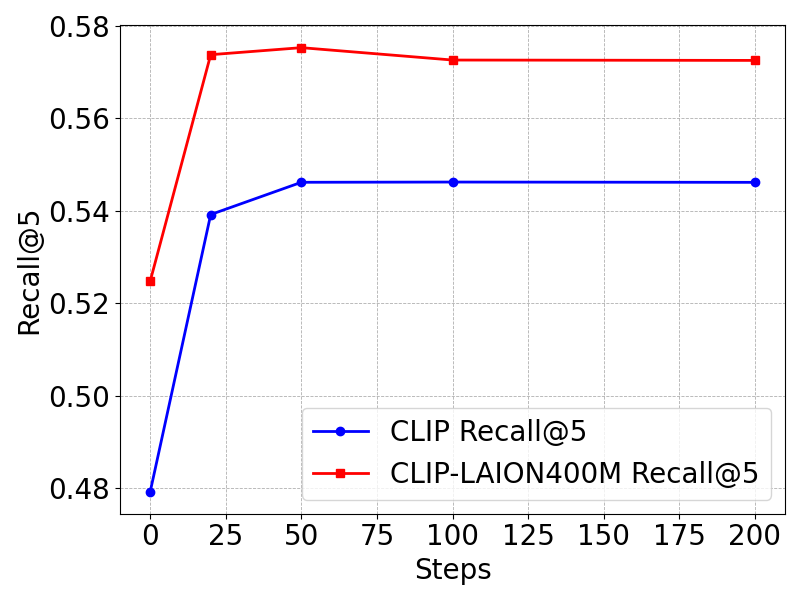}
    \caption{\textbf{Retrieval performance with respect to the number of optimization steps on COCO-Neg}.}
\label{figure:opt_image}
\vspace{-0.4cm}
\end{figure}

\subsubsection{Effect of Query Decomposition}
To analyze the role of query decomposition, we conduct an ablation study on the decomposition step. Table~\ref{ablation_decompose} reports results on the NegConstraint dataset with BGE-M3. We introduce an Only Decompose variant that retrieves using the averaged embedding of decomposed positive and negative sub-queries, and an Only Decompose (RRF) variant that retrieves separately for each sub-query and merges results using Reciprocal Rank Fusion (RRF) with k=60. As shown in Table~\ref{ablation_decompose}, query decomposition alone yields only limited improvements over the baseline. While RRF-based aggregation provides additional gains, the performance remains far below that of the full method. This indicates that the performance improvements do not stem from decomposition itself, but from the subsequent embedding optimization.

\begin{figure}[t!] 
\includegraphics[width=\linewidth]{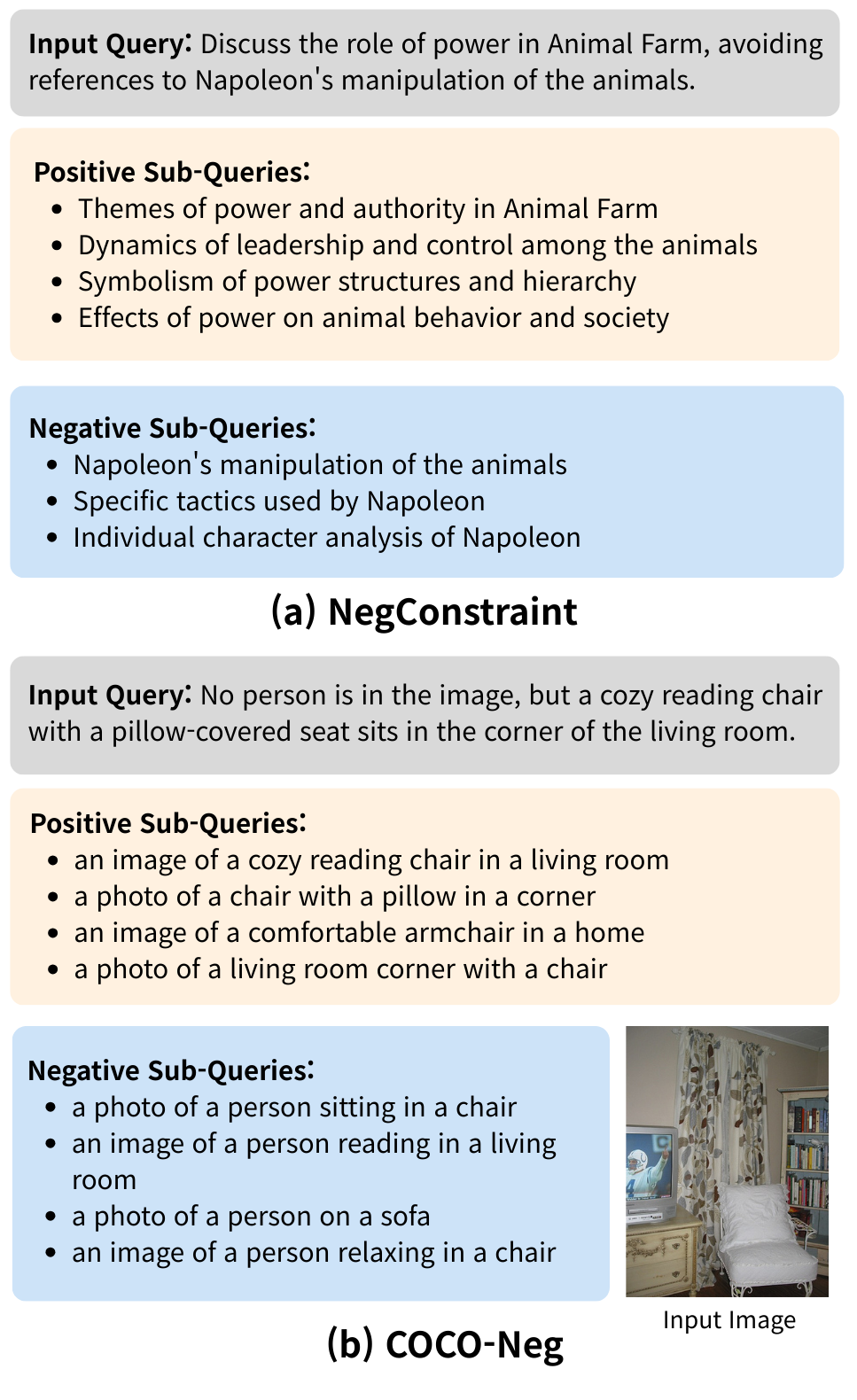}
\caption{\textbf{Examples of query decomposition.} In NegConstraint (a) and COCO-Neg (b), the LLM decomposes queries into positive sub-queries capturing desired elements and negative sub-queries targeting excluded concepts.}
\label{figure:sample}
\vspace{-0.4cm}
\end{figure}

\subsubsection{Effect of Optimization Steps}
We analyze the effect of the number of optimization steps on retrieval performance. On NegConstraint (Figure~\ref{figure:opt_text}), performance improves sharply when increasing steps from 0 to 20, and remains stable between 20 and 50 steps. However, beyond 100 steps, both nDCG@10 and MAP@100 gradually decline. On COCO-Neg
(Figure~\ref{figure:opt_image}), Recall@5 similarly improves from 0
to 20 steps, peaks around 50 steps, and slightly decreases beyond 100 steps. In both cases, 20 to 50 steps is sufficient to achieve strong performance, and we adopt 20 steps as the default setting across all experiments.

\section{Analysis}
\subsection{Quality of Query Decomposition}
We analyze how the LLM decomposes queries into positive and negative sub-queries. To evaluate decomposition quality, we perform a binary correctness assessment on NegConstraint~\cite{xu2025logical} using GPT-4.1-mini, measuring whether each output captures the intended positive and negative components of the original query. The decomposition achieves 91.76\% accuracy, indicating that generated sub-queries are largely aligned with user intent. In addition, we empirically verify that the input queries are effectively decomposed on both NegConstraint and COCO-Neg~\cite{alhamoud2025vision}, as illustrated in Figure~\ref{figure:sample}.

\begin{figure}[t!] 
\includegraphics[width=\linewidth]{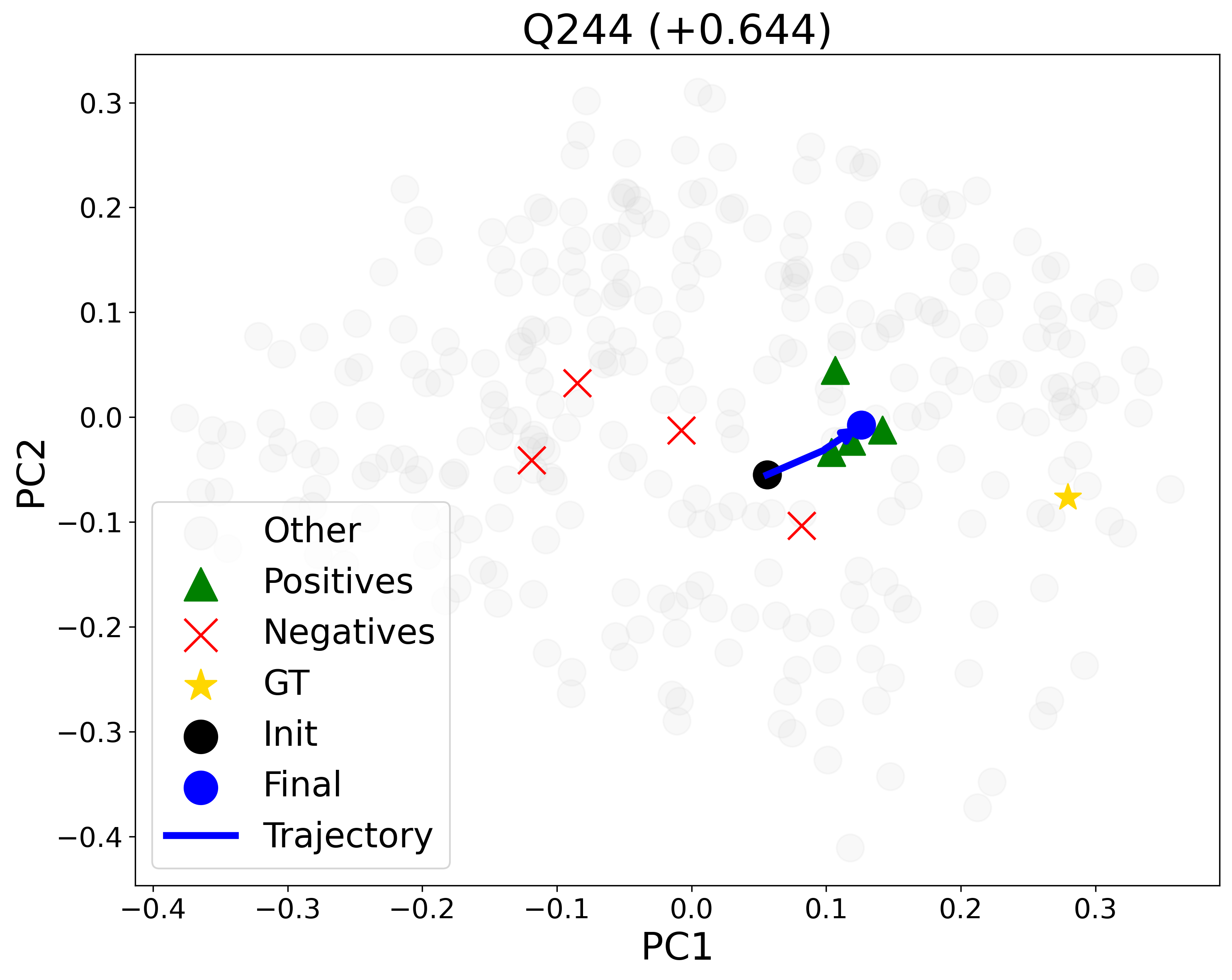}
\caption{\textbf{Trajectory of optimized query embedding $\mathbf{e}_u$ in PCA-projected space.} 
The initial embedding (black $\bullet$) moves toward the final state (blue $\bullet$). 
Positive examples (green $\triangle$) and the ground-truth (yellow $\star$) act as attractors, 
while negative examples (red $\times$) exert repelling forces. Other corpus embeddings are shown in light gray.}
    \label{fig:trajectory}
\end{figure}

\subsection{Embedding Space Analysis}
We visualize how the optimized query embedding $\mathbf{e}_u$ moves relative to positives ($\mathbf{e}_p$), negatives ($\mathbf{e}_n$), and corpus embeddings. We fit a PCA on the full corpus embeddings of BGE-M3 and project the query trajectory, decomposed sub-queries, and ground-truth documents onto the same 2D space. Experiments are conducted on NegConstraint~\cite{xu2025logical} with the same setting as Sec~\ref{sec:exp_setup}.

Figure~\ref{fig:trajectory} illustrates a representative example where the query asks for \textit{journalistic evaluation criteria while excluding their sources and the Pulitzer Prize for History}. The initial embedding (black) progressively shifts toward the final optimized point (blue), attracted by positive sub-queries (green triangles) and the gold document (yellow star), while repelled from negative sub-queries (red crosses). In the baseline, a document about the \textit{Pulitzer Prize for History}---which should be excluded---ranks 1st, pushing the ground-truth to rank 6. After optimization, the ground-truth rises to rank 1 (nDCG@10: 0.356$\rightarrow$1.0). Across the top-5 improved queries, ground-truth documents move from ranks 6--33 into the top 2, with an average nDCG@10 gain of +0.63. This confirms that the contrastive optimization effectively reshapes $\mathbf{e}_u$ toward relevant regions while suppressing negatives.

\begin{figure}[t!] 
\includegraphics[width=\linewidth]{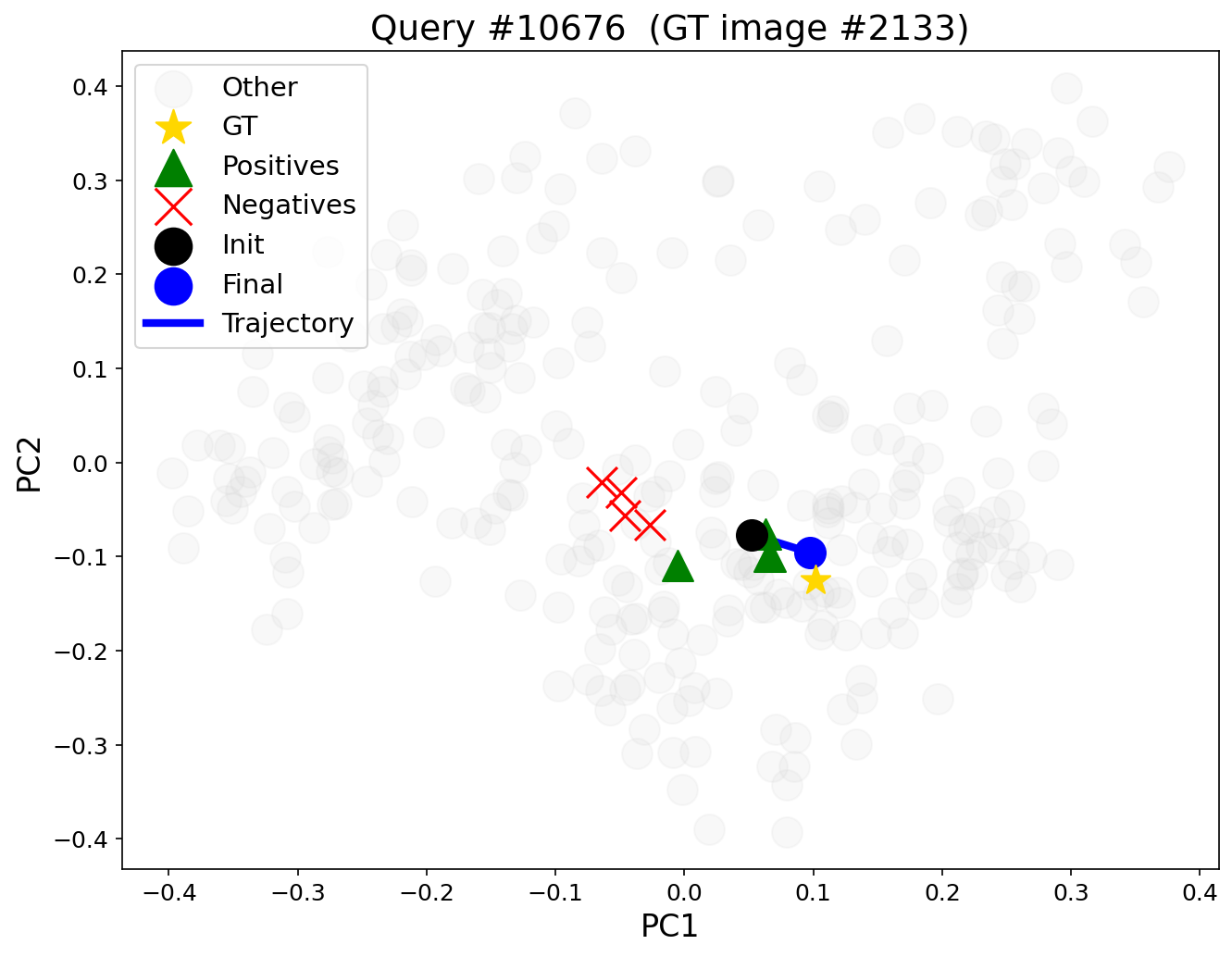}
\caption{\textbf{Optimization trajectory of $\mathbf{e}_u$ in PCA-projected CLIP space.} 
Initial (black $\bullet$) embedding evolves toward final state (blue $\bullet$), 
attracted by positives (green $\triangle$) and ground-truth (yellow $\star$) 
while repelled by negatives (red $\times$). Gray points denote the corpus.}
    \label{fig:trajectory_coco}
    \vspace{-0.6cm}
\end{figure}

We further extend this analysis to the vision-language setting. Using CLIP ViT-B/32 on the COCO negated retrieval benchmark~\cite{alhamoud2025vision}, we fit PCA on the CLIP image embeddings and project the text query trajectory onto the same 2D space. Figure~\ref{fig:trajectory_coco} shows that the same trajectory pattern holds: the initial text embedding (black) shifts toward the ground-truth image (yellow star) and positive sub-queries (green triangles), while moving away from negative sub-queries (red crosses) that encode the negated concept. This demonstrates that DEO generalizes across both text retrieval and cross-modal image retrieval, effectively reshaping the query embedding in CLIP's shared vision-language space to suppress negated attributes.

\subsection{Computational Efficiency}
To evaluate the computational overhead of our parameter optimization, we measured runtime on both CPU and GPU environments. On a CPU (AMD Ryzen 7 5800X 8-Core Processor, 64.0GB RAM), DEO with 20 optimization steps required a total of 0.016 seconds (average 0.000665 seconds per step), while 50 steps required 0.035 seconds (average 0.000640 seconds per step). On a GPU (NVIDIA GeForce RTX 3060 12GB), DEO with 20 optimization steps required a total of 0.033 seconds (average 0.00172 seconds per step), while 50 steps required 0.095 seconds (average 0.001932 seconds per step). These results demonstrate that our method remains highly efficient across both CPU and GPU settings, making it practical for real-world applications where GPU resources may be limited or additional training data is unavailable.
\section{Conclusion}
In this work, we propose Direct Embedding Optimization (DEO), a simple yet effective method for negation- and exclusion-aware retrieval that does not require fine-tuning or additional datasets. By decomposing user queries into positive and negative sub-queries and directly optimizing the embedding space with contrastive loss, DEO aligns query embeddings more precisely with user intent. Our experiments demonstrate that DEO consistently outperforms baseline methods across both text and multimodal retrieval tasks, achieving substantial gains on benchmarks involving negation and exclusion. These results highlight the robustness and practicality of our approach in real-world scenarios where negation and exclusion are frequently present in user queries.

\section{Limitations}
While DEO proves effective without fine-tuning, it relies on the ability of LLMs to correctly decompose user queries into positive and negative sub-queries. As shown in Sec~\ref{sec:ablation}, the final retrieval performance may vary depending on the decomposition quality of the LLM. We believe that DEO provides a promising direction for building lightweight and controllable retrieval systems. Future work could explore enhancing query decomposition with more robust LLMs, incorporating adaptive optimization strategies that automatically select loss balancing parameters per query, and extending DEO to diverse multimodal datasets beyond images, such as audio.

\section*{Acknowledgments}
This work was supported by Miri.DIH.

\bibliography{custom}





\end{document}